\documentclass[twoside,11pt]{article}

\usepackage{jmlr2e}
\usepackage[caption=false]{subfig}
\usepackage{upquote}

\usepackage{xcolor}
\newcommand{\greencheck}{{\color{green}\checkmark}}

\newcommand{\redx}{{\color{red}X}}

\usepackage{booktabs}

\jmlrheading{19}{2018}{1-5}{08/18}{10/18}{burns18a}{David M. Burns and Cari M. Whyne}

\ShortHeadings{Seglearn: Learning Sequences and Time Series}{Burns and Whyne}
\firstpageno{1}

\begin{document}

\title{Seglearn: A Python Package for Learning Sequences and Time Series}

\author{\name David M. Burns \email d.burns@utoronto.ca \\       
       \addr Sunnybrook Research Institute \\
       \AND
       \name Cari M. Whyne \email cari.whyne@sunnybrook.ca \\
       \addr Sunnybrook Research Institute \\
       2075 Bayview Ave. Room S620.\\
       Toronto, ON, Canada. M4N 3M5. \\ 
       }

\editor{Alexandre Gramfort}

\maketitle

\begin{abstract}
\texttt{seglearn} is an open-source Python package for performing machine learning on time series or sequences. The implementation provides a flexible pipeline for tackling classification, regression, and forecasting problems with multivariate sequence and contextual data. Sequences and series may be learned directly with deep learning models or via feature representation with classical machine learning estimators. This package is compatible with \texttt{scikit-learn} and is listed under \texttt{scikit-learn} "Related Projects". The package depends on \texttt{numpy}, \texttt{scipy}, and \texttt{scikit-learn}. \texttt{seglearn} is distributed under the BSD 3-Clause License. Documentation includes a detailed API description, user guide, and examples. Unit tests provide a high degree of code coverage. Source code and documentation can be downloaded from \texttt{https://github.com/dmbee/seglearn}. 
\end{abstract}


\section{Introduction}

Many real-world machine learning problems e.g. voice recognition, human activity recognition, power systems fault detection, stock price and temperature prediction, involve data that is captured as sequences over a period of time \citep{aha_uci_2018}. Sequential data sets do not fit the standard supervised learning framework, where each sample $(\mathbf{x},y)$ within the data set is assumed to be independently and identically distributed (iid) from a joint distribution $P(\mathbf{x},y)$ \citep{bishop_pattern_2011}. Instead, the data consist \textit{sequences} of $(\mathbf{x},y)$ pairs, and nearby values of $(\mathbf{x},y)$ within a \textit{sequence} are likely to be correlated to each other. Sequence learning exploits the sequential relationships in the data to improve algorithm performance.

\section{Supported Problem Classes}

Sequence data sets have a general formulation \citep{dietterich_machine_2002} as sequence pairs $\{(\mathbf{X}_i,\mathbf{y}_i)\}_{i=1}^{N}$, where each $\mathbf{X}_i$ is a multivariate sequence with $T_i$ samples $\langle \mathbf{x}_{i,1}, \mathbf{x}_{i,2},...,\mathbf{x}_{i,T_i} \rangle$ and each $\mathbf{y}_i$ target is a univariate sequence with $T_i$ samples $ \langle y_{i,1}, y_{i,2},..., y_{i,T_i} \rangle $. The targets $\mathbf{y}_i$ can either be sequences of categorical class labels (for classification problems), or sequences of continuous data (for regression problems). The number of samples $T_i$ varies between the sequence pairs in the data set. Time series with a regular sampling period may be treated equivalently to sequences. Irregularly sampled time series are formulated with an additional sequence variable $\mathbf{t}_i$ that increases monotonically and indicates the timing of samples in the data set $\{(\mathbf{t}_i, \mathbf{X}_i,\mathbf{y}_i)\}_{i=1}^{N}$.  

\paragraph{}
Important sub-classes of the general sequence learning problem are sequence classification and sequence prediction. In sequence classification problems (eg song genre classification), the target for each sequence is a fixed class label $y_i$ and the data takes the form $\{(\mathbf{X}_i, y_i)\}_{i=1}^{N}$. Sequence prediction involves predicting a future value of the target $(y_{i,t+f})$ or future values $ \langle y_{i,t+1}, y_{i,t+2},..., y_{i,t+f} \rangle $, given
$\langle \mathbf{x}_{i,1}, \mathbf{x}_{i,2},...,\mathbf{x}_{i,t} \rangle$, $ \langle y_{i,1}, y_{i,2},..., y_{i,t} \rangle $, and sometimes also $\langle \mathbf{x}_{i,t+1}, \mathbf{x}_{i,t+2},...,\mathbf{x}_{i,t+f} \rangle$.

\paragraph{}
A final important generalization is the case where contextual data associated with each sequence, but not varying within the sequence, exists to support the machine learning algorithm performance. Perhaps the algorithm for reading electrocardiograms will be given access to laboratory data, the patient's age, or known medical diagnoses to assist with classifying the sequential data recovered from the leads. 

\paragraph{}
\texttt{seglearn} provides a flexible, user-friendly framework for learning time series and sequences in all of the above contexts. Transforms for sequence padding, truncation, and sliding window segmentation are implemented to fix sample number across all sequences in the data set. This permits utilization of many classical and modern machine learning algorithms that require fixed length inputs. Sliding window segmentation transforms the sequence data into a piecewise representation (segments), which is particularly effective for learning periodized sequences \citep{bulling_tutorial_2014}. An interpolation transform is implemented for resampling irregularly sampled time series. The sequence or time series data can be learned directly with various neural network architectures \citep{lipton_critical_2015}, or via a feature representation which greatly enhances performance of classical algorithms \citep{bulling_tutorial_2014}. 

\section{Installation}

The \texttt{seglearn} source code is available at: \texttt{https://github.com/dmbee/seglearn}. It is operating system agnostic, and implemented purely in Python. The dependencies are \texttt{numpy}, \texttt{scipy}, and \texttt{scikit-learn}. The package can be installed using pip:  \\
\texttt{\$ pip install seglearn} 
\paragraph{}
Alternatively, seglearn can be installed from the sources: \\
\texttt{\$ git clone https://github.com/dmbee/seglearn} \\
\texttt{\$ cd seglearn} \\
\texttt{\$ pip install .}
\paragraph{}
Unit tests can be run from the root directory using \texttt{pytest}.

\section{Implementation}

\begin{figure}[!htb]
  \begin{center}
    \includegraphics[scale=0.5]{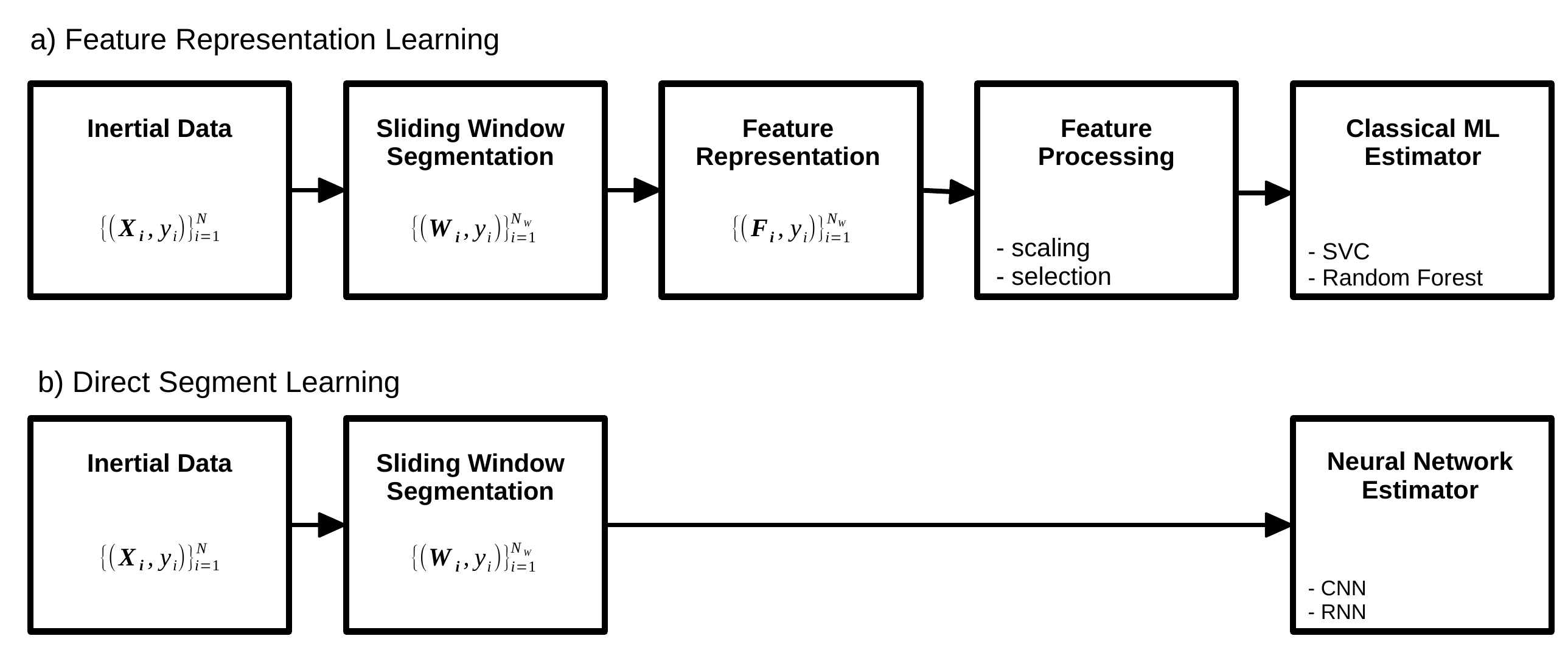}
  \end{center}
  \caption{Example \texttt{seglearn} pipelines for a) learning segment feature representations, b) learning segments directly. SVC: Support Vector Classifier, CNN: Convolution Neural Network, RNN: Recurrent Neural Network.}
  \label{f:pipe}
\end{figure}

The \texttt{seglearn} API was implemented for compatibility with \texttt{scikit-learn} and its existing framework for model evaluation and selection. The \texttt{seglearn} package provides means for handling sequence data, segmenting it, computing feature representations, calculating train-test splits and cross-validation folds along the temporal axis.\footnote{Note splitting time series data along the temporal axis violates the assumption of independence between train and test samples. However, this is useful in some cases, such as the analysis of a single series.} An iterable, indexable data structure is implemented to represent sequence data with supporting contextual data. 

\paragraph{}
The \texttt{seglearn} functionality is provided within a \texttt{scikit-learn} pipeline allowing the user to leverage \texttt{scikit-learn} transformer and estimator classes, which are particularly helpful in the feature representation approach to segment learning. Direct segment learning with neural networks is implemented in pipeline using the \texttt{keras} package, and its \texttt{scikit-learn} API. Examples of both approaches are provided in the documentation and example gallery. The integrated learning pipeline, from raw data to final estimator, can be optimized within the \texttt{scikit-learn model{\_}selection} framework. This is important because segmentation parameters (eg window size, segment overlap) can have a significant impact on sequence learning performance \citep{burns_shoulder_2018, bulling_tutorial_2014}. 

\paragraph{}
Sliding window segmentation transforms sequence data into a piecewise representation (segments), such that predictions are made and scored for all segments in the data set. Sliding window segmentation can be performed for data sets with a single target value per sequence, in which case that target value is mapped to all segments generated from the parent sequence. If the target for each is sequence is also a sequence, the target is segmented as well and various methods may be used to select a single target value from the target segment (e.g. mean value, middle value, last value, etc.) or the target segment sequence can be predicted directly if an estimator implementing sequence to sequence prediction is utilized.

\paragraph{}
A human activity recognition data set \citep{burns_shoulder_2018} consisting of inertial sensor data recorded by a smartwatch worn during shoulder rehabilitation exercises is provided with the source code to demonstrate the features and usage of the \texttt{seglearn} package. 

\section{Basic Example}

This example demonstrates the use of \texttt{seglearn} for performing sequence classification with our smartwatch human activity recognition data set.
\texttt{\\
$>>>$ import seglearn as sgl \\
$>>>$ from sklearn.model{\_}selection import train{\_}test{\_}split \\
$>>>$ from sklearn.ensemble import RandomForestClassifier \\
$>>>$ from sklearn.preprocessing import StandardScaler \\
$>>>$ \\
$>>>$ data = sgl.load{\_}watch() \\
$>>>$ X{\_}train, X{\_}test, y{\_}train, y{\_}test = train{\_}test{\_}split(data["X"], data["y"]) \\
$>>>$ \\
$>>>$ clf = sgl.Pype([("seg", sgl.SegmentX(width=100, overlap=0.5)), \\
...\hspace{106pt}("features", sgl.FeatureRep()), \\
...\hspace{106pt}("scaler", StandardScaler()), \\
...\hspace{106pt}("rf", RandomForestClassifier())]) \\
$>>>$ \\
$>>>$ clf.fit(X{\_}train, y{\_}train) \\
$>>>$ score = clf.score(X{\_}test, y{\_}test) \\
$>>>$ print("accuracy score:", score) \\\\
accuracy score: 0.7805084745762711 \\}

\section{Comparison to other Software}
Three other Python packages for performing machine learning on time series and sequences were identified: \texttt{tslearn} \citep{tavenard_tslearn:_2017}, \texttt{cesium-ml} \citep{naul_cesium:_2016}, and \texttt{tsfresh} \citep{christ_time_2018}. These were compared to \texttt{seglearn} based on time series learning capabilities (Table \ref{t1:features}), and performance (Table \ref{t2:perf}). 

\texttt{cesium-ml} (v0.9.6) and \texttt{tsfresh} (v0.11.1) support feature representation learning of multi-variate time series, and currently implement more features than does \texttt{seglearn}. However, the feature representation transformers are implemented as a pre-processing step, independent to the otherwise sklearn compatible pipeline. This design choice precludes end-to-end model selection. There are no examples or apparent support for problems where the target is a sequence/time series or integration with deep learning models. 

\texttt{tslearn} (v0.1.18.4) implements time-series specific classical algorithms for clustering, classification, and barycenter computation for time series with varying lengths. There is no support for feature representation learning, learning context data, or deep learning. 

The performance comparison was conducted using our human activity recognition data set with 140 multivariate time series with 6 channels sampled uniformly at 50 Hz and 7 activity classes. The series' were all truncated to 4 seconds (200 samples). Classification accuracy was measured on 35 series' held out for testing, and 105 used for training. \texttt{seglearn}, \texttt{cesium-ml}, and \texttt{tsfresh} were tested using the sklearn implementation of the SVM classifier with a radial basis function (RBF) kernel on 5 features (median, minimum, maximum, standard deviation, and skewness) calculated on each channel (total 30 features). \texttt{tslearn} was evaluated with its own SVM classifier implementing a global alignment kernel \citep{cuturi_kernel_2007}. The testing was performed using an Intel Core i7-4770 testbed with 16 GB of installed memory, on Linux Mint 18.3 with Python 2.7.12. 

Classification accuracy was identical between \texttt{cesium-ml}, \texttt{tsfresh}, and \texttt{seglearn} (as they used the same features and classifier in the evaluation), and all three significantly exceeded the accuracy achieved with \texttt{tslearn}. \texttt{seglearn} significantly outperformed the other packages in terms of computation time. 
\begin{table}[]
	\centering
	\begin{tabular}{lcccc}
		\toprule
		& \textbf{tslearn} & \textbf{cesium-ml} & \textbf{ts-fresh} & \textbf{seglearn} \\ \midrule
		Active development (2018)          & \greencheck              & \greencheck                & \greencheck               & \greencheck               \\
		Documentation                      & \greencheck              & \greencheck                & \greencheck               & \greencheck               \\
		Unit Tests                         & \greencheck              & \greencheck                & \greencheck               & \greencheck               \\
		Multivariate time series           & \greencheck              & \greencheck                & \greencheck               & \greencheck               \\
		Context data                       & \redx               & \greencheck                & \redx                & \greencheck               \\
		Time series target                 & \redx               & \redx                 & \redx                & \greencheck               \\
		Sliding window segmentation        & \redx               & \redx                 & \redx                & \greencheck               \\
		Temporal folds                     & \redx               & \redx                 & \redx                & \greencheck               \\
		sklearn compatible model selection & \redx               & \redx                 & \redx                & \greencheck               \\
		Feature representation learning    & \redx               & \greencheck                & \greencheck               & \greencheck               \\
		Number of implemented features     & N/A              & 58                 & 64                & 20                \\
		Deep learning                      & \redx               & \redx                 & \redx                & \greencheck               \\
		Classification                     & \greencheck              & \greencheck                & \greencheck               & \greencheck               \\
		Clustering                         & \greencheck              & \greencheck                & \greencheck               & \greencheck               \\
		Regression                         & \greencheck              & \greencheck                & \greencheck               & \greencheck               \\
		Forecasting                        & \redx               & \greencheck                & \greencheck               & \greencheck              \\ \bottomrule
	\end{tabular}
	\caption{Comparison of time series learning package features for \texttt{tslearn} v0.1.18.4, \texttt{cesium-ml} v0.9.6, \texttt{tsfresh} v0.11.1 and \texttt{seglearn} v1.0.2.}
	\label{t1:features}
\end{table}

\begin{table}[]
	\centering
	\begin{tabular}{@{}lcccc@{}}
		\toprule
		& \texttt{tslearn} & \texttt{cesium-ml} & \texttt{ts-fresh} & \texttt{seglearn}        \\ \midrule
		Classification accuracy & 0.057 & \textbf{0.714} & \textbf{0.714}  & \textbf{0.714} \\
		Computation time (seconds) & 0.79 & 62.9 & 0.40 & \textbf{0.088} \\ \bottomrule
	\end{tabular}
	\caption{Comparison of time series learning package performance on our human activity recognition dataset.}
	\label{t2:perf}
\end{table}

\newpage

\vskip 0.2in
\bibliography{burns18a}

\begin{thebibliography}{10}
\providecommand{\natexlab}[1]{#1}
\providecommand{\url}[1]{\texttt{#1}}
\expandafter\ifx\csname urlstyle\endcsname\relax
  \providecommand{\doi}[1]{doi: #1}\else
  \providecommand{\doi}{doi: \begingroup \urlstyle{rm}\Url}\fi

\bibitem[Aha(2018)]{aha_uci_2018}
David Aha.
\newblock {UCI} {Machine} {Learning} {Repository}, March 2018.
\newblock URL \url{https://archive.ics.uci.edu/ml/index.php}.

\bibitem[Bishop(2011)]{bishop_pattern_2011}
Christopher~M. Bishop.
\newblock \emph{Pattern {Recognition} and {Machine} {Learning}}.
\newblock Springer, New York, 2nd edition, April 2011.
\newblock ISBN 978-0-387-31073-2.

\bibitem[Bulling et~al.(2014)Bulling, Blanke, and
  Schiele]{bulling_tutorial_2014}
Andreas Bulling, Ulf Blanke, and Bernt Schiele.
\newblock A tutorial on human activity recognition using body-worn inertial
  sensors.
\newblock \emph{ACM Computing Surveys}, 46\penalty0 (3):\penalty0 1--33,
  January 2014.
\newblock ISSN 03600300.
\newblock \doi{10.1145/2499621}.

\bibitem[Burns et~al.(2018)Burns, Leung, Hardisty, Whyne, Henry, and
  McLachlin]{burns_shoulder_2018}
David Burns, Nathan Leung, Michael Hardisty, Cari Whyne, Patrick Henry, and
  Stewart McLachlin.
\newblock Shoulder {Physiotherapy} {Exercise} {Recognition}: {Machine}
  {Learning} the {Inertial} {Signals} from a {Smartwatch}.
\newblock \emph{arXiv:1802.01489 [cs]}, February 2018.
\newblock arXiv: 1802.01489.

\bibitem[Christ et~al.(2018)Christ, Braun, Neuffer, and
  Kempa-Liehr]{christ_time_2018}
Maximilian Christ, Nils Braun, Julius Neuffer, and Andreas~W. Kempa-Liehr.
\newblock Time {Series} {FeatuRe} {Extraction} on basis of {Scalable}
  {Hypothesis} tests (tsfresh – {A} {Python} package).
\newblock \emph{Neurocomputing}, 307:\penalty0 72--77, September 2018.
\newblock ISSN 0925-2312.
\newblock \doi{10.1016/j.neucom.2018.03.067}.
\newblock URL
  \url{http://www.sciencedirect.com/science/article/pii/S0925231218304843}.

\bibitem[Cuturi et~al.(2007)Cuturi, Vert, Birkenes, and
  Matsui]{cuturi_kernel_2007}
Marco Cuturi, Jean-Philippe Vert, Oystein Birkenes, and Tomoko Matsui.
\newblock A {Kernel} for {Time} {Series} {Based} on {Global} {Alignments}.
\newblock In \emph{2007 {IEEE} {International} {Conference} on {Acoustics},
  {Speech} and {Signal} {Processing} - {ICASSP} '07}, pages II--413--II--416,
  Honolulu, HI, April 2007. IEEE.
\newblock ISBN 978-1-4244-0727-9.
\newblock \doi{10.1109/ICASSP.2007.366260}.
\newblock URL \url{http://ieeexplore.ieee.org/document/4217433/}.

\bibitem[Dietterich(2002)]{dietterich_machine_2002}
Thomas~G. Dietterich.
\newblock Machine {Learning} for {Sequential} {Data}: {A} {Review}.
\newblock In \emph{Structural, {Syntactic}, and {Statistical} {Pattern}
  {Recognition}}. Springer, Berlin, Heidelberg, 2002.
\newblock ISBN 978-3-540-44011-6 978-3-540-70659-5.
\newblock \doi{10.1007/3-540-70659-3_2}.

\bibitem[Lipton et~al.(2015)Lipton, Berkowitz, and Elkan]{lipton_critical_2015}
Zachary~C. Lipton, John Berkowitz, and Charles Elkan.
\newblock A critical review of recurrent neural networks for sequence learning.
\newblock \emph{arXiv preprint arXiv:1506.00019}, 2015.

\bibitem[Naul et~al.(2016)Naul, van~der Walt, Crellin-Quick, Bloom, and
  Pérez]{naul_cesium:_2016}
Brett Naul, Stéfan van~der Walt, Arien Crellin-Quick, Joshua~S. Bloom, and
  Fernando Pérez.
\newblock cesium: {Open}-{Source} {Platform} for {Time}-{Series} {Inference}.
\newblock \emph{arXiv:1609.04504 [cs]}, September 2016.
\newblock arXiv: 1609.04504.

\bibitem[Tavenard(2017)]{tavenard_tslearn:_2017}
Romain Tavenard.
\newblock tslearn: {A} machine learning toolkit dedicated to time-series data,
  2017.
\newblock URL \url{https://github.com/rtavenar/tslearn}.

\end{thebibliography}

\end{document}